\PassOptionsToPackage{numbers}{natbib}
\documentclass{article}

\usepackage[preprint]{neurips_2025}  
\usepackage[numbers]{natbib}

\usepackage{scalefnt,letltxmacro}
\usepackage[acronym,nowarn]{glossaries}
\glsdisablehyper
\makeglossaries
\usepackage{xspace}
\usepackage{float}
\usepackage{physics}
\usepackage[normalem]{ulem}
\usepackage[english]{babel}

\usepackage{enumitem} 
\usepackage{booktabs} 
\usepackage{array}    
\usepackage{wrapfig}  
\usepackage{lipsum}   
\usepackage[]{microtype} 

\usepackage{amssymb}
\usepackage{amsmath}
\usepackage{amsfonts}
\usepackage{mathtools}
\usepackage{dsfont} 

\usepackage[dvipsnames,svgnames,table]{xcolor}
\usepackage[colorlinks]{hyperref}
\definecolor{pink3}{HTML}{fc0557}
\definecolor{shadecolor}{gray}{0.9}
\definecolor{mylightgray}{gray}{0.94}
\usepackage[all]{hypcap}
\hypersetup{
  citecolor=pink3,
  linkcolor=pink3,
  urlcolor=pink3
}
\usepackage[most,skins,theorems]{tcolorbox}
\definecolor{MyGreen}{RGB}{34, 139, 34}
\newcommand{\improve}[1]{\textcolor{MyGreen}{$\blacktriangle${\small #1}\%}}
\newcommand{\improvesmall}[1]{\textcolor{MyGreen}{$\blacktriangle${\footnotesize #1}\%}}
\usepackage[capitalize,nameinlink]{cleveref}
\crefname{section}{\S}{\S\S}
\Crefname{section}{\S}{\S\S}

\usepackage{graphicx}
\usepackage{subcaption}
\usepackage{tikz}
\usetikzlibrary{shapes.geometric}
\usepackage[export]{adjustbox}

\usepackage{pgfplots}
\pgfplotsset{compat=1.6}
\usepgfplotslibrary{groupplots}
\pgfplotsset{
  tick label style = {font=\sffamily},
  every axis label/.append style={font=\sffamily},
  typeset ticklabels with strut,
  xticklabel={$\mathsf{\pgfmathprintnumber{\tick}}$},
  yticklabel={$\mathsf{\pgfmathprintnumber{\tick}}$},
  every axis title/.append style={yshift=-1ex}
}

\usepackage[colorinlistoftodos,
  textsize=scriptsize,
  linecolor=BurntOrange,
  bordercolor=BurntOrange,
  backgroundcolor=BurntOrange!25]{todonotes}

\usepackage{multirow}
\usepackage{pifont}

\newcommand{\cmark}{\ding{51}}
\newcommand{\xmark}{\ding{55}}

\usepackage{url}

\usepackage[normalem]{ulem} 

\tcbset{
  promptstyle/.style={
      enhanced,
      colback=white,
      colframe=black,
      colbacktitle=gray!20,
      width=0.98\linewidth,
      coltitle=black,
      rounded corners,
      sharp corners=north,
      boxrule=0.5pt,
      drop shadow=black!50!white,
      attach boxed title to top left={
          xshift=-2mm,
          yshift=-2mm
        },
      title code={%
          \begin{minipage}{0.8\linewidth}
            \raggedright\thetitle
          \end{minipage}
        },
      boxed title style={
          rounded corners,
          size=small,
          colback=gray!20,
        },
      fonttitle=\normalsize\bfseries,
      before upper={\parindent15pt\raggedright}
    }
}

\tcbset{
  aibox/.style={
      width=\linewidth,
      top=8pt,
      bottom=4pt,
      colback=blue!6!white,
      colframe=black,
      colbacktitle=black,
      enhanced,
      center,
      attach boxed title to top left={yshift=-0.1in,xshift=0.15in},
      boxed title style={boxrule=0pt,colframe=white,},
    }
}
\newtcolorbox{AIbox}[2][]{aibox,title=#2,#1}

\usepackage{listings}
\lstdefinestyle{TinyJSON}{
morekeywords={
    _question_, _evidence_, _database_schema_, im_start, im_end, user, system, assistant, <answer>, think, Wait,
  },
basicstyle=\ttfamily\footnotesize, 
keywordstyle=\color{purple}\bfseries, 
stringstyle=\color{purple}, 
commentstyle=\color{gray}\itshape, 
numbers=none, 
numberstyle=\tiny\color{gray}, 
stepnumber=1, 
numbersep=10pt, 
tabsize=2, 
captionpos=b, 
breaklines=true, 
breakatwhitespace=true, 
showspaces=false, 
showstringspaces=false, 
escapeinside={(*@}{@*)}, 
frame=none, 
extendedchars=true,
literate={á}{{\'a}}1 {ã}{{\~a}}1 {é}{{\'e}}1 {í}{{\'i}}1 {ó}{{\'o}}1 {ú}{{\'u}}1 {ç}{{\c{c}}}1,
}

\definecolor{mygreen}{HTML}{008000}
\lstdefinestyle{SQL}{
  language=SQL,
  morekeywords={
      SELECT, RETURN, WHERE, CREATE, DELETE, DETACH, SET, MERGE, ON,
      OPTIONAL, WITH, DISTINCT, AS, LIMIT, ORDER, BY, SKIP, ASC, DESC,
      AND, OR, NOT, IN, IS, STARTS, ENDS, CONTAINS, TRUE, FALSE, NULL, CALL, UNWIND, FROM, GROUP, BY, HAVING, AVG, MAX, MIN,
      CASE, WHEN, THEN, ELSE, END, JOIN, INNER, LEFT, RIGHT, FULL, CROSS, UNION, INTERSECT, EXCEPT
    },
  basicstyle=\ttfamily\small, 
  keywordstyle=\color{mygreen}\bfseries,
  stringstyle=\color{purple},
  commentstyle=\color{gray}\itshape,
  numbers=left, 
  numberstyle=\tiny\color{gray},
  stepnumber=1,
  numbersep=10pt,
  tabsize=2,
  captionpos=b,
  breaklines=true,
  breakatwhitespace=true,
  showspaces=false,
  showstringspaces=false,
  escapeinside={(*@}{@*)},
  frame=single, 
  framesep=5pt, 
  aboveskip=5pt, 
  belowskip=5pt, 
  extendedchars=true,
}

\newcommand{\mathbold}[1]{{\boldsymbol{#1}}}
\newcommand{\nestedmathbold}[1]{{\mathbold{#1}}}

\newcommand{\mbs}{\nestedmathbold{s}}

\newcommand{\mbx}{\nestedmathbold{x}}
\newcommand{\mby}{\nestedmathbold{y}}
\newcommand{\mbz}{\nestedmathbold{z}}

\newcommand{\mbtheta}{\nestedmathbold{\theta}}

\newcommand{\E}{\mathbb{E}}

\newcommand{\bbR}{\mathbb{R}}

\DeclareRobustCommand{\KL}[2]{\textsc{kl}\left[#1\;\|\;#2\right]}

\newcommand{\cD}{\mathcal{D}}

\newcommand{\cL}{\mathcal{L}}

\newcommand{\cM}{\mathcal{M}}
\newcommand{\cP}{\mathcal{P}}

\newcommand{\cS}{\mathcal{S}}
\newcommand{\cT}{\mathcal{T}}

\newcommand{\cX}{\mathcal{X}}

\newacronym{SGD}{SGD}{stochastic gradient descent}
\newacronym{SGA}{SGA}{stochastic gradient ascent}
\newacronym{MAP}{MAP}{maximum-a-posteriori}
\newacronym{MLE}{MLE}{maximum likelihood estimation}
\newacronym{MNLL}{MNLL}{mean negative log-likelihood}
\newacronym{NLL}{NLL}{negative log-likelihood}
\newacronym{LL}{LL}{log-likelihood}
\newacronym{RMSE}{RMSE}{root mean square error}
\newacronym{ECE}{ECE}{expected calibration error}
\newacronym{SNR}{SNR}{signal-to-noise ratio}
\newacronym{FID}{FID}{Fr\'echet Inception Distance}
\newacronym{BPD}{BPD}{bit per dimension}
\newacronym{NFE}{NFE}{neural function evaluations}

\newacronym{AE}{AE}{autoencoder}
\newacronym{WAE}{WAE}{Wasserstein Autoencoder}
\newacronym{VAE}{VAE}{Variational Autoencoder}
\newacronym{BAE}{BAE}{Bayesian autoencoder}
\newacronym{GAN}{GAN}{Generative Adversarial Network}
\newacronym{DPGMM}{DPGMM}{Dirichlet process Gaussian mixture model}

\newacronym{MC}{MC}{Monte Carlo}
\newacronym{MCMC}{MCMC}{Markov chain Monte Carlo}
\newacronym{HMC}{HMC}{Hamiltonian Monte Carlo}
\newacronym{MH}{MH}{Metropolis-Hastings}
\newacronym{NUTS}{NUTS}{no-u-turn sampler}
\newacronym{SGHMC}{SGHMC}{stochastic gradient Hamiltonian Monte Carlo}
\newacronym{OU}{OU}{Ornstein-Uhlenbeck}

\newacronym{SDE}{SDE}{Stochastic Differential Equation}
\newacronym{CNF}{CNF}{Continuous Normalizing Flow}
\newacronym{ODE}{ODE}{Ordinary Differential Equation}
\newacronym{NF}{NF}{normalizing flow}

\newacronym{GP}{GP}{Gaussian Process}
\newacronym[longplural=deep Gaussian processes]{DGP}{DGP}{deep Gaussian process}
\newacronym{GPLVM}{GPLVM}{Gaussian process latent variable model}
\newacronym{DPMM}{DPMM}{Dirichlet Process Mixture Model}

\newacronym{LLM}{LLM}{large language model}
\newacronym{VI}{VI}{variational inference}
\newacronym{SVI}{SVI}{stochastic variational inference}
\newacronym{BNN}{BNN}{Bayesian neural network}
\newacronym{DNN}{DNN}{deep neural network}
\newacronym{CNN}{CNN}{convolutional neural network}
\newacronym{MLP}{MLP}{multilayer perceptron}
\newacronym{NN}{NN}{neural network}
\newacronym{RELU}{ReLU}{rectified linear unit}

\newacronym{ELBO}{ELBO}{evidence lower bound}
\newacronym{NELBO}{NELBO}{negative evidence lower bound}
\newacronym{ELL}{ELL}{expected log likelihood}
\newacronym{KL}{KL}{Kullback-Leibler divergence}
\newacronym{AUC}{AUC}{area under the curve}
\newacronym{VFE}{VFE}{variational free energy}

\newacronym{RBF}{RBF}{radial basis function}
\newacronym{ARD}{ARD}{automatic relevance determination}
\newacronym{RKHS}{RKHS}{reproducing kernel Hilbert space}

\newacronym{OT}{OT}{optimal transport}
\newacronym{WD}{WD}{Wasserstein distance}
\newacronym{SWD}{SWD}{sliced-Wasserstein distance}
\newacronym{DSWD}{DSWD}{distributional sliced-Wasserstein distance}

\newacronym{LAP}{LAP}{linear assignment problem}
\newacronym{SOLAP}{SOLAP}{sum of bilinear assignment problems}

\newacronym{ICL}{ICL}{in-context learning}
\newacronym{LoRA}{LoRA}{Low-Rank Adaptation}
\newacronym{SFT}{SFT}{Supervised Fine-Tuning}
\newacronym{RL}{RL}{reinforcement learning}
\newacronym{RLHF}{RLHF}{reinforcement learning from human feedback}
\newacronym{PPO}{PPO}{Proximal Policy Optimization}
\newacronym{GRPO}{GRPO}{Grouped Proximal Policy Optimization}
\newacronym{COT}{CoT}{chain-of-thought}

\usepackage[hyperpageref]{backref}
\renewcommand*{\backrefalt}[4]{%
\ifcase #1 %
No citations.%
\or
(p. #2)%
\else
(pp. #2)%
\fi
}

\newcommand{\stitle}[1]{\vspace{1ex}\noindent{\textbf{#1}}}

\title{
Think2SQL: Reinforce LLM Reasoning Capabilities for Text2SQL
}

\author{%
  Simone Papicchio \\
  Politecnico di Torino, Turin, Italy\\
  EURECOM, Biot, France \\
  \texttt{simone.papicchio@polito.it} \\
  \And
  Simone Rossi \\
  EURECOM, Biot, France \\
  \texttt{simone.rossi@eurecom.fr} \\
  \And
  Luca Cagliero \\
  Politecnico di Torino, Turin, Italy\\
  \texttt{luca.cagliero@polito.it} \\
  \And
  Paolo Papotti \\
  EURECOM, Biot, France \\
  \texttt{paolo.papotti@eurecom.fr}\\
}

\begin{document}

\maketitle

\begin{abstract}
  Large Language Models (LLMs) can translate natural language into SQL, but small models struggle with multi-table and complex queries in Zero-Shot Learning (ZSL) settings. While Supervised Fine-Tuning (SFT) helps, it falls short for harder cases. To address this, we study how different reasoning strategies—general-purpose reasoning in ZSL, reasoning traces in SFT, and Reinforcement Learning with Verifiable Reward (RLVR) with novel reward functions—affect Text2SQL performance across four benchmarks. We show that partial scoring rewards, computed via SQL execution, are crucial for guiding models even when outputs are not fully correct. These fine-grained signals lead to consistently better Text2SQL outcomes. Small LLMs benefit most from reasoning-aware SFT and RL, with the 14B Qwen-Coder-2.5 surpassing 400B+ models on challenging datasets like BIRD.
\end{abstract}



\begin{figure}[htbp]
    \centering
    \includegraphics[width=0.6\textwidth]{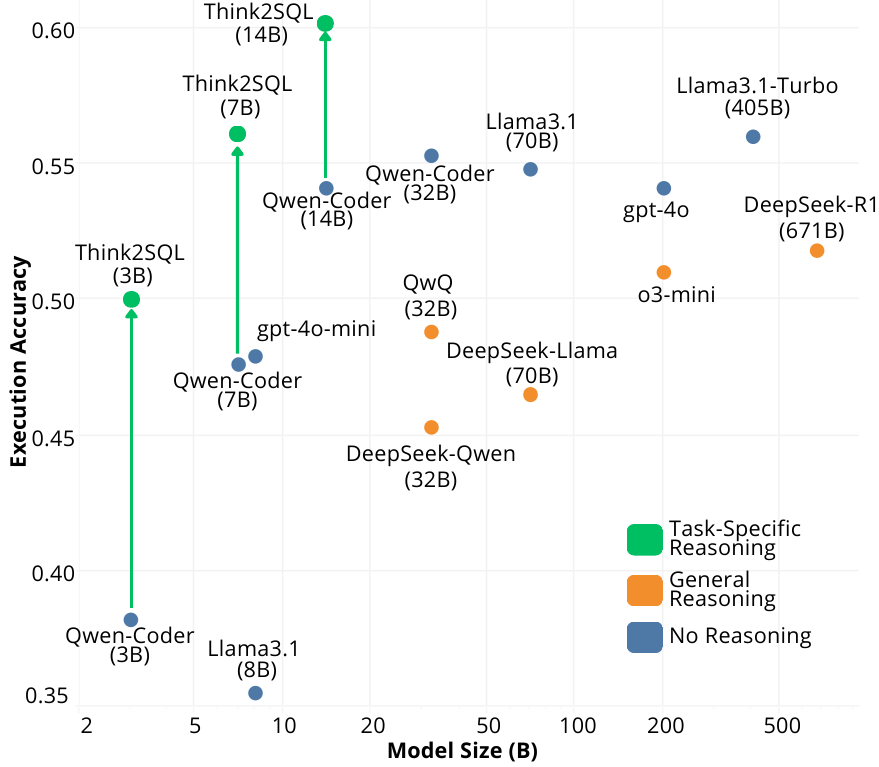}
    \caption{\textbf{Execution accuracy vs. model size (log scale) for various open- and closed- models on the BIRD development set.} Orange points indicate models with general reasoning, blue points models without reasoning and green points task-specific reasoning. Vertical green lines show performance gains given by RLVR only with the novel rewards for the same base model. All models are instructed versions. \texttt{Qwen-Coder} is the 2.5 version.}
    \label{fig:size_vs_performance}
\end{figure}

\section{Introduction}
The exponential growth of data stored in relational databases, coupled with the rapid advancements of Large Language Models (LLMs) have jointly paved the way for new accessible ways to query large multi-table databases.
This task, known as Text-to-SQL (or Text2SQL), allows users to interact with databases by converting natural language questions about relational tables into executable SQL queries~\cite{10.14778/3681954.3681960}.

%
%
Thanks to the advanced language understanding capabilities of pretrained models, LLMs have remarkably improved the Text2SQL performance of traditional neural network-based solutions (e.g. \cite{XiaoDG16,BoginBG19,tacl_survey}), particularly on multi-table datasets~\cite{li2024bird,yu2018spider}.
However, tackling Text2SQL effectively remains a particularly challenging problem not only because it involves expressing first- or second-order logic conditions in SQL but also because it requires reasoning about the underlying question's meaning and its relation to the database schema~\cite{Floratou2024NL2SQLIA}.
Additionally, LLMs' performance under Zero-Shot Learning (ZSL) significantly varies depending on the number of model parameters~\cite{10.1145/3641289} and the training data used for pretraining~\cite{abs-2409-02038}.
Large models, albeit powerful, lack domain specificity, whereas smaller models suffer from limited reasoning depth~\cite{Survey-Text2SQL-LLMs}.
%
%

To improve performance, LLMs can be adapted to the Text2SQL task through various training methods.
Supervised Fine-Tuning (SFT) is a widely adopted adaptation strategy that involves training LLMs on annotated question-SQL pairs~\cite{WeiBZGYLDDL22,10.1145/3654930}.
While SFT can enhance performance, it requires a large amount of task-specific data, which is often difficult to obtain~\cite{10.1145/3654930}.
Furthermore, smaller LLMs still show limited generality and logical inference on complex SQL~\cite{papicchio2025qatch}.
Reinforcement Learning with Verifiable Reward (RLVR) has emerged as a promising approach to enhance LLM reasoning capabilities~\citep{lambert2024t,deepseekai2025deepseekr1}.
Unlike SFT, in RLVR Text2SQL is casted as a decision-making problem, where a model generates SQL queries in response to natural language inputs and learns from rule-based signals.
The model, treated as an agent, receives rewards based on how well its output matches the intended query.
These rewards can be derived from execution accuracy~\cite{yu2018spider,li2024bird} or more detailed evaluations of query structure and content~\cite{papicchio2023qatch}.
Through iterative interactions, the model updates its parameters to favor actions (e.g. SQL queries) that yield higher rewards, enabling improved reasoning and generalization beyond supervised examples.
While RLVR has shown success in various domains, including mathematical reasoning and code generation, its application to Text2SQL remains unexplored~\cite{NGUYEN2025100135,zhai2025excot,pourreza2025reasoning}.

In this paper, we investigate the effect of LLMs' reasoning capabilities on Text2SQL performance through a comparative analysis under distinct training configurations, with a particular focus on the effectiveness and adaptability of small LLMs.
In particular, we aim to answer the following central question: \textit{To what extent do reasoning capabilities embedded or learned by LLMs enhance Text2SQL performance across different training paradigms?}

We find that augmenting small and mid-size LLMs with \emph{task-specific reasoning traces} during supervised fine-tuning boosts execution accuracy by up to 10 absolute points, and that subsequent reinforcement learning with a \emph{dense, partial-credit} reward delivers a further 15–30 \% relative gain as shown in Figure~\ref{fig:size_vs_performance}. This combination raises a 7B model to an accuracy on the BIRD benchmark that surpasses open-weight 70B models and proprietary \texttt{GPT-4o}. We observe similar advantages on other benchmarks, 
demonstrating that 
reasoning supervision can outweigh 
orders of magnitude in scale.

\textbf{Contributions.}
In order to address the aforementioned research question,
(1) we evaluate the impact of reasoning on Text2SQL across multiple LLM training regimes, including Zero-Shot, SFT, RLVR and their combination;
(2) we introduce a novel reasoning-augmented SFT strategy that leverages SQL queries enriched with logical traces to guide model training;
(3) we propose a RLVR approach that leverages rule-based rewards~\cite{deepseekai2025deepseekr1}. These rewards are derived from both standard execution accuracy metrics~\cite{yu2018spider} and 
fine-grained instance-level metrics, which evaluate precision, recall, and cardinality of partially correct answers~\cite{papicchio2023qatch};
(4) we compare the generalization trade-offs of small versus large LLMs on diverse Text2SQL benchmarks;
(5) we provide an empirical analysis on how reasoning integration affects performance across query complexity and domain variability.

\section{Methodology}
\label{sec:preliminaries}
To evaluate the influence of reasoning on the Text2SQL task,
we evaluate the impact of reasoning on Text2SQL across multiple LLM training regimes.
Text2SQL was chosen for its practical importance, recent advancements~\cite{Google2025Gemini}, and LLM familiarity. In addition, SQL queries are executable and verifiable, making them ideal for this study.

This section is structured as follows: (i) we define the Text2SQL problem and its formalization, (ii) we detail the Supervised Fine-Tuning (SFT) training procedure and the dataset creation, and (iii) we outline the Reinforcement Learning with Verifiable Reward (RLVR) training framework.

\subsection{Text2SQL problem formulation}
Text2SQL translates a natural language question $\mbx$ into an SQL query $\mby$ based on a database schema $\cS$, which defines tables, attributes, and data types. Auxiliary context $\cM$, such as metadata or prior query examples, may also be included.

Let $\pi_{\mbtheta}$ denote a Large Language Model (LLM) parameterized by $\mbtheta$, formalized as a probabilistic autoregressive decoder-only model~\cite{radford2018improving}. The model $\pi_{\mbtheta}$ defines a conditional distribution over the SQL query $\mby = (y_1, \ldots, y_T)$ given the input $\mbx$ and the schema context:
\begin{equation}
  \label{eq:text2sql}
  \pi_{\mbtheta}(\mby \mid \mbx, \cS, \cM) = \prod_{t=1}^{T} \pi_{\mbtheta}(y_t \mid \mbx \Vert \cS \Vert \cM \Vert \mby_{<t})\,,
\end{equation}
where $\Vert$ denotes the concatenation operator. We adopt the prompt format widely used in prior studies for schema representation due to its demonstrated effectiveness in Text2SQL tasks~\cite{chang2023prompt,gao2023text}. Supervised Fine-Tuning (SFT) details are in~\cref{sec:finetuning}.


Modern Text2SQL systems restrict $\cS$ to a localized substructure $\cS_{\mbx}$,
where $\cS_{\mbx} \subseteq \cS$ is retrieved via schema linking, lexical overlap, or
learned attention~\cite{wang2019ratSql,pourreza2023din,talaei2024chessSql,chen2024tablerag,caferouglu2024sql}.
To better isolate the reasoning process in Text2SQL and disentangle schema linking from SQL generation,
we restrict to $\cS_{\mbx}$ during training and inference. This does not affect the generality of our approach, as modern Text2SQL systems can integrate multi-table retrievers~\cite{wu2025mmqa} to fetch relevant schema information dynamically.

\subsection{Supervised Fine-Tuning (SFT) and distilled dataset}
\label{sec:sft}

The SFT approach trains the model, as formalized in \cref{eq:text2sql}, to maximize the likelihood associated with next-token prediction. This sequence consists of the system prompt, user prompt, and distilled reasoning trace with the corresponding answer. We use the system prompt from DeepSeek-R1, and our reasoning traces are also collected using this model~\cite{deepseekai2025deepseekr1}.
An example of training prompt is in~\cref{sec:prompt-training}. The loss function is computed exclusively on the distilled reasoning traces and answers to optimize training efficiency, ignoring input tokens~\citep{chiang2023vicuna,yu-etal-2024-lions}. This is particularly beneficial as the distilled traces are significantly longer than the input sequence~\cite{shi2024instruction}. Additional details on the SFT objective can be found in~\cref{sec:finetuning}.

To construct our distilled SFT dataset, we start with the BIRD dataset~\cite{li2024bird}, which comprises 9,428 examples spanning 69 databases across 37 professional domains, including healthcare, blockchain, and education. Each example consists of a \textit{natural language (NL)} question, a corresponding \textit{SQL query}, and supplementary \textit{evidence} to address ambiguities in the schema or question. We employed a rigorous two-step curation process~\cite{muennighoff2025s1simpletesttimescaling}: (1) filtering out erroneous SQL queries and duplicates, resulting in the removal of 421 entries; and (2) categorizing the remaining queries by complexity—\emph{Simple} ($[0,7)$), \emph{Medium} ($[7,10)$), and \emph{Challenging} ($[10,+\infty)$)—based on the number of SQL constructs. This process yielded 6,404 simple queries (70.6\%), 2,140 medium-complexity queries (23.6\%), and 530 challenging queries (5.8\%). The complexity intervals were determined using the distribution observed in the BIRD development sets, where complexity annotations are available.

We annotated a total of 1,142 examples using the DeepSeek-R1 model~\cite{deepseekai2025deepseekr1}, leveraging its official system prompt and reasoning-optimized hyperparameters (temperature=0.7, top-p=0.95) \citep{togheterai2025promptDeepseek,openaiReasoning2024}, as detailed in \cref{sec:prompt-synthetic-data-annotation}. To construct a high-quality dataset with reliable reasoning traces, we began with the most challenging examples and retained the 193 instances for which the model achieved an execution accuracy of 1.0. We then randomly sampled 500 medium-complexity and 800 simple-complexity queries, from which the model correctly generated 265 and 684 valid answers, respectively. This sampling strategy ensures a balanced coverage across complexity tiers while maintaining reduced cost~\cite{muennighoff2025s1simpletesttimescaling}.  The 75th percentile of reasoning token lengths for the retained examples is 509 for simple, 861 for medium, and 869 for challenging queries.



\begin{figure}
    \centering
    \includegraphics[width=1\textwidth]{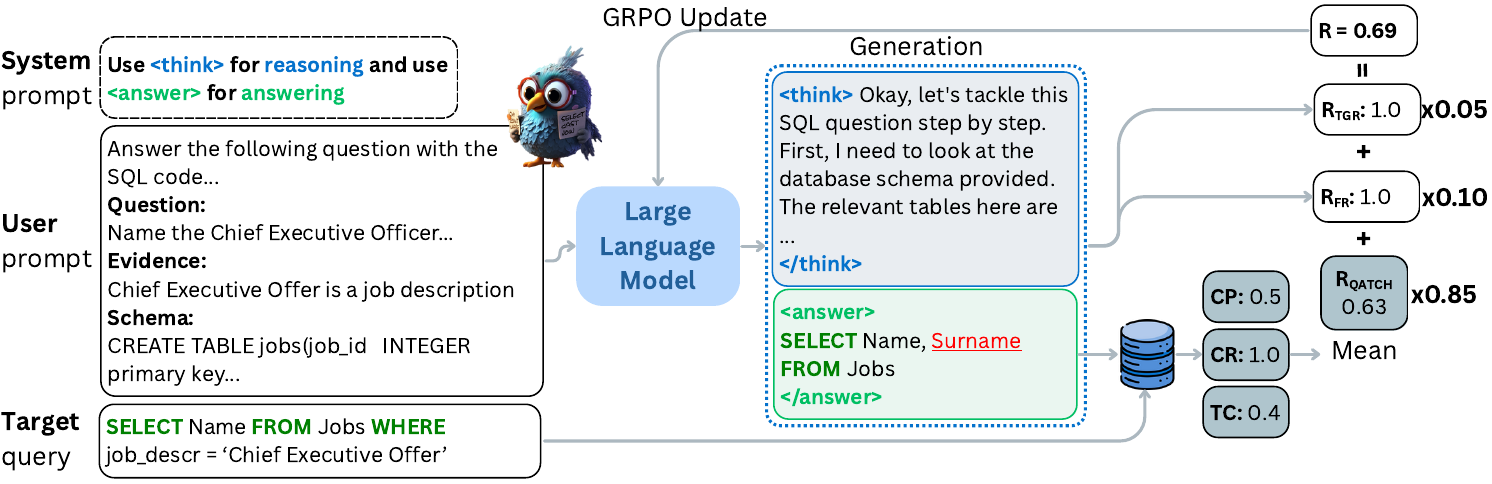}
    \caption{\textbf{Overview of the RLVR algorithm.} An LLM—either a base or SFT model—generates multiple outputs per step, each comprising a reasoning trace and a predicted SQL query. Rewards are computed as a weighted sum of format adherence and SQL correctness, incorporating metrics such as \textit{Cell Precision} (CP), \textit{Cell Recall} (CR), and \textit{Tuple Cardinality} (TC)~\cite{papicchio2025qatch}. These rewards guide model updates via GRPO.}
    \label{fig:methodology_overview}
\end{figure}

\subsection{Reinforcement Learning with Verifiable Reward (RLVR)}
\label{sec:rlvr}

Reinforcement Learning with Verifiable Reward (RLVR)~\citep{lambert2024t,deepseekai2025deepseekr1,team2025kimi} extends the RLHF~\cite{ouyang2022rlhf} framework by replacing the learned reward model with a rule-based, verifiable reward. This reduces reward hacking and spurious correlations~\citep{gao2023rewardhack,everitt2021rewardhack}, and has been shown to enhance specific cognitive behaviors~\citep{gandhi2025cognitive}. Despite reward sparsity—where feedback is only provided upon strict constraint satisfaction—RLVR is effective when initialized from a strong pretrained model and applied to tasks with verifiable objectives, such as code or SQL generation~\citep{le2022coderl,gehring2024rlef,chen2023teaching}.

Training is performed using Group-Relative Policy Optimization (GRPO)~\cite{shao2024deepseekmath}, detailed in~\cref{sec:grpo}. As shown in~\cref{fig:methodology_overview}, the model takes a prompt (\textit{question}, \textit{evidence}, and \textit{database schema}, see~\cref{sec:prompt-training}) and generates multiple outputs, each with a reasoning trace and SQL query. A weighted reward—assessing format compliance and SQL correctness—is computed and used to update the model. In the following, we refer to task-specific RLVR for the models trained with both the reward function and the reasoning traces designed for the Text2SQL task. 

\stitle{Reward Design.}
The first reward function we can think of is the execution accuracy (EX). This binary metric evaluates whether the generated SQL query execution matches the target SQL execution row by row. While straightforward, this reward can result in sparse feedback.
To address this limitation, we adopt a dense reward function that provides partial credit for partially correct SQL queries. This is achieved by incorporating the QATCH metrics~\citep{papicchio2023qatch,papicchio2025qatch}, which 
assign a score between 0 and 1, reflecting the degree of correctness of the generated SQL query. For example, if the predicted SQL query includes extra columns or misses some required columns, the metrics adjust the reward accordingly. This approach ensures the model receives feedback for partially correct SQL queries, encouraging incremental improvements during training, especially for small LLMs. Additionally, we introduce two auxiliary signals to ensure compliance with the expected output structure~\cite{deepseekai2025deepseekr1}. 
Mathematical definitions of these rewards are provided in~\cref{sec:rewards}, with a summary as follows:

\underline{\textit{Execution Accuracy (EX) Reward.}} Defined as $R_{\text{EX}}\in\{0,1\}$ is a binary reward indicating whether the generated SQL query execution matches the target SQL execution row by row.

\underline{\textit{QATCH Reward.}} Defined as $R_{\text{QATCH}} \in [0, 1]$, this score is obtained as the average of three sub-metrics {computed by executing the queries}
~\citep{papicchio2023qatch,papicchio2025qatch}. Considering the example in~\cref{fig:methodology_overview}:
(i) Cell Precision (CP) measures the fraction of correctly projected columns. In the example, $CP = 0.5$ because the predicted SQL includes an extra column—\textit{Surname}—not present in the target.
(ii) Cell Recall (CR) measures the fraction of target columns that are present in the predicted SQL. In the example, $CR = 1$ since all required target columns are included in the prediction.
(iii) Tuple Cardinality (TC) measures the ratio of output tuples between the predicted and target SQL. In the example, $TC = 0.4$ because the predicted SQL lacks a filtering condition, resulting in a larger result set than the target SQL.

\underline{\textit{Format Reward (FR).}} Defined as $R_{\text{FR}}$ is a binary reward indicating whether the output matches the expected format, 
defined as a reasoning trace (in the \verb|<think>| and \verb|</think>| tags) followed by a SQL query (enclosed within \verb|<answer>| and \verb|</answer>| tags).

\underline{\textit{Tag Count Reward (TCR).}} Defined as $R_{\text{TCR}} \in \{0, 0.25, 0.5, 0.75, 1\}$ is a reward that penalizes the overuse or omission of tags. It assigns a score of 0.25 for each tag that appears exactly once in the output. The tags considered are \verb|<think>|, \verb|</think>|, \verb|<answer>|, and \verb|</answer>|.

The final reward
is computed as a weighted sum of the individual rewards. Let $R_{\text{text2SQL}}$ be $R_{\text{EX}}$ or $R_{\text{QATCH}}$. The final reward is defined as:
\[
    R = 0.85 \cdot R_{\text{text2SQL}} + 0.10 \cdot R_{\text{FR}} + 0.05 \cdot R_{\text{TCR}}, \quad R \in [0,1]
\]

When R$_{\text{EX}} = 1$, R$_{\text{QATCH}}$ is also 1, so combining the rewards adds no value. Since QATCH is only informative when R$_{\text{EX}} = 0$, sharing weights between them limits its impact. Empirically, this combination leads to suboptimal performance. Therefore, we keep them separate and defer joint analysis to future work.

\begin{table}
    \small
    \centering
    \setlength{\tabcolsep}{5pt}
    \caption{\textbf{Performance comparison of open-source and proprietary models on the BIRD Dev dataset. }
        All models were evaluated with a temperature setting of 0.7 and a top\_p value of 0.95.
        \texttt{Llama} models correspond to version 3.1 and Turbo means the model is quantized 8bit.
        Think2SQL models denote the Qwen2.5-Coder models trained exclusively with RLVR and $R_{\text{QATCH}}$.}
    \label{tab:bird-dev-analysis}
    \begin{tabular}{l@{\hskip 0.1pt}c|ccc|c}
        \toprule
        \textbf{Model}                  & \textbf{Reasoning} & \textbf{Simple} & \textbf{Medium} & \textbf{Challenging} & \textbf{Weigthed AVG} \\
        \midrule
        \rowcolor[gray]{0.9}
        \multicolumn{6}{l}{\textit{Open-source LLMs ($<$ 10B)}} \vspace{2pt}                                                                    \\
        \texttt{DeepSeek-Qwen-1.5B}     & \cmark             & $0.056$         & $0.004$         & $0.0$                & $0.035$               \\
        \texttt{Qwen2.5-Coder-0.5B}     & \cmark             & $0.126 $        & $0.033$         & $0.035$              & $0.089$               \\
        \texttt{DeepSeek-Qwen-7B}       & \cmark             & $0.297$         & $0.113$         & $0.049$              & $0.218$               \\
        \texttt{Qwen2.5-Coder-1.5B}     & \cmark             & $0.351$         & $0.184$         & $0.077$              & $0.275$               \\
        \texttt{Llama-8b}               & \xmark             & $0.436$         & $0.260$         & $0.133$              & $0.355$               \\
        \texttt{Qwen2.5-Coder-3B}       & \xmark             & $0.469$         & $0.267$         & $0.196$              & $0.382$               \\
        \texttt{Qwen2.5-Coder-7B}       & \xmark             & $0.548$         & $0.388$         & $0.294$              & $\textbf{0.476}$      \\
        \rowcolor[gray]{0.9}
        \multicolumn{6}{l}{\textit{Open-source LLMs (10-100B)}} \vspace{2pt}                                                                    \\
        \texttt{DeepSeek-Qwen-32B}      & \cmark             & $0.542$         & $0.347$         & $0.217$              & $0.453$               \\
        \texttt{DeepSeek-Llama-70B}     & \cmark             & $0.552$         & $0.371$         & $0.203$              & $0.465$               \\
        \texttt{QwQ-32B}                & \cmark             & $0.550$         & $0.427$         & $0.280$              & $0.488$               \\
        \texttt{Qwen2.5-Coder-14B}      & \xmark             & $0.610$         & $0.456$         & $0.364$              & $0.541$               \\
        \texttt{Llama-70B}              & \xmark             & $0.618$         & $0.469$         & $0.350$              & $0.548$               \\
        \texttt{Qwen2.5-Coder-32B}      & \xmark             & $0.623$        & $0.482$         & $0.329$              & $\textbf{0.553}$      \\
        \rowcolor[gray]{0.9}
        \multicolumn{6}{l}{\textit{Open-source LLMs ( $>$100B)}} \vspace{2pt}                                                                   \\
        \texttt{DeepSeek-R1}            & \cmark             & $0.588$         & $0.440$         & $0.294$              & $0.518$               \\
        \texttt{Llama-405B-Turbo}       & \xmark             & $0.630$         & $0.477$         & $0.371$              & $\textbf{0.560}$      \\
        \rowcolor[gray]{0.9}
        \multicolumn{6}{l}{\textit{Closed-source LLMs}} \vspace{2pt}                                                                            \\
        \texttt{gpt-4o-mini-2024-07-18} & \xmark             & $0.545$         & $0.401$         & $0.301$              & $0.479$               \\
        \texttt{o3-mini-2025-01-31}     & \cmark             & $0.561$         & $0.406$         & $0.329$              & $0.510$               \\
        \texttt{gpt-4o-2024-08-06}      & \xmark             & $0.619$         & $0.447$         & $0.343$              & $\textbf{0.541}$      \\
        \midrule
        \rowcolor[gray]{0.9}
        \multicolumn{6}{l}{\textit{Our Models}} \vspace{2pt}                                                                                    \\
        \texttt{Think2SQL-0.5B}           & \cmark             &      0.342    &   0.122       & 0.086              & 0.254               \\
        \texttt{Think2SQL-1.5B}           & \cmark             & 0.526         & 0.333        & 0.226             & 0.442               \\
        \texttt{Think2SQL-3B}           & \cmark             & $0.574$         & $0.403$         & $0.336$              & $0.500$               \\
        \texttt{Think2SQL-7B}           & \cmark             & $0.628$         & $0.482$         & $0.385$              & 0.561                 \\
        \texttt{Think2SQL-14B}          & \cmark             & 0.654           & 0.543           & 0.462                & \textbf{0.602}        \\
    \end{tabular}
\end{table}

\section{Experiments}
\label{sec:exp}
In this work, we analyze how reasoning capabilities affect the performance of small LLMs on the Text2SQL task. Our central question is: \textit{To what extent do reasoning capabilities embedded or learned by LLMs enhance Text2SQL performance across different training paradigms?} 
While various techniques exist to push results in Text2SQL~\cite{Chung2025IsLC}, our goal is not to compete on leaderboard metrics. Instead, we systematically compare models—with and without explicit reasoning components—under different training configurations to isolate and understand the contribution of reasoning to overall performance.

\subsection{Experiments Setup}
\stitle{Training setup.}
We use the \texttt{Qwen-2.5-Coder-Instruct} family trained with Open-R1~\citep{openr1,vonwerra2022trl} and a fixed seed of 42. SFT models (suffix \texttt{SFT}) are trained for 5 epochs on 4×A100 80GB GPUs (batch size 128, learning rate $4\times10^{-5}$, AdamW~\cite{loshchilov2017decoupled}) using the distilled dataset of 1,142 stratified samples (913 train, 229 val), taking 10 minutes (3B) and 20 minutes (7B). RLVR models (suffix \texttt{RL}) are trained with GRPO~\cite{shao2024deepseekmath} for 1 epoch (batch size 256, micro-batch 64, learning rate $1\times10^{-6}$, 16 generations/batch) on 8×H100 80GB GPUs (4 update, 4 generation) over 40h (3B) and 60h (7B), using 9,007 BIRD samples (567 gradient steps) and VLLM~\cite{kwon2023efficient} with FlashAttention~\cite{dao2022flashattention}, temperature 0.7, top-$p$ 0.95 max generation token 4096. An LRU cache optimizes reward computation. Cold-start models (\texttt{SFT-RL}) follow DeepSeek-R1~\cite{deepseekai2025deepseekr1}, with similar training time to RLVR. All models use a common prompt (\cref{sec:prompt-training}) and restrict schema inputs to relevant tables. Best-performing models on BIRD dev are denoted \texttt{Think2SQL-3B}, \texttt{Think2SQL-7B}, and \texttt{Think2SQL-14B}.

\stitle{Model Baselines.}
We evaluate both open- and closed-source models, with and without reasoning, in a zero-shot setting using a shared prompt (\cref{sec:prompt-training}), temperature 0.7, top-$p$ 0.95, 30k token limit, and seed 42. Each experiment is run 3 times; we report the mean (std $<$ 0.01 omitted). Inference is performed with LightEval~\cite{lighteval} via VLLM, configured as in RLVR. Closed-source models (e.g., \texttt{GPT-4o}, \texttt{o3-mini}~\cite{o3_mini_system_card,openai_gpt4o_system_card}) are accessed through LiteLLM~\cite{litellm2025}, and open-source models $>$70B via Together-AI~\cite{togetherai2025}.
We assess two main families: \texttt{Qwen-Coder-2.5} and \texttt{Llama} 3.1~\cite{grattafiori2024llama3}. \texttt{Qwen-Coder} enables direct comparison between (our) reasoning and non-reasoning variants, including \texttt{QwQ}~\cite{qwq32b}. \texttt{Llama} models benchmark \texttt{DeepSeek-R1} distilled versions (e.g., 405B \texttt{Llama} and 671B \texttt{DeepSeek-R1}). All models are instruction-tuned.





\stitle{Evaluation.} We evaluate on the BIRD development set (1,530 instances: 924 simple, 461 medium, 143 challenging), as the test set is unavailable. To assess robustness and generalization, we also use SPIDER~\cite{yu2018spider} and its variants: Spider-Syn~\cite{gan2021spidersyn}, which tests robustness to schema-related paraphrasing, and Spider-DK~\cite{gan2021spiderdk}, which introduces implicit domain knowledge. Additionally, we include KaggleDBQA~\cite{lee-etal-2021-kaggledbqa}, a challenging cross-domain dataset featuring real-world unnormalized schemas, domain-specific data types, and natural language questions that reflect realistic user queries.

\subsection{Main Results}
\begin{wrapfigure}{r}{0.27\textwidth} 
    \centering
    \vspace{-2.8ex}
    \includegraphics[width=\linewidth]{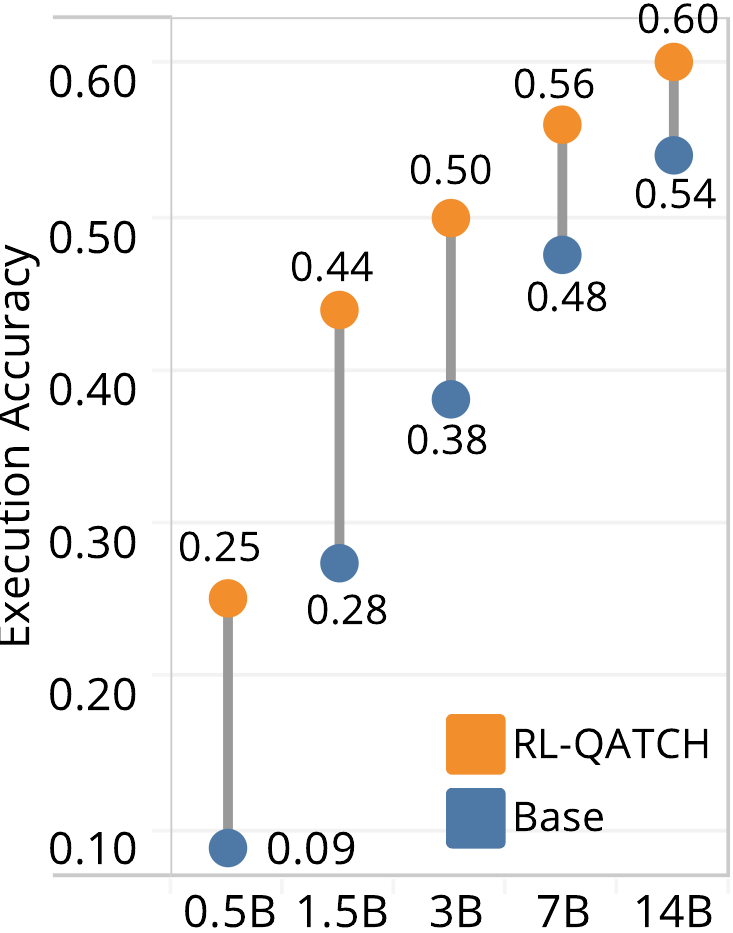}
    \vspace{-3ex}
    \caption{\textbf{\texttt{Qwen-Coder} size vs. EX. on BIRD Dev.}}
    \label{fig:model_size_ex_vs_ex}
    \vspace{-2.5ex}
\end{wrapfigure}
\cref{tab:bird-dev-analysis} presents a performance comparison of various open-source and proprietary models on the BIRD Dev dataset for the Text2SQL task, categorized by model size and reasoning capability. Among open-source models, the \texttt{Qwen2.5-Coder-32B} achieved the highest weighted average accuracy (0.553) in the 10-100B category, while \texttt{Llama-405B-Turbo} led the $>$100B category with a weighted average of 0.560. Proprietary models also performed competitively, with \texttt{gpt-4o-2024-08-06} achieving a weighted average of 0.541.  The Think2SQL models demonstrate strong performance across all evaluation
categories, particularly on challenging examples.

\texttt{Think2SQL-14B} achieves a weighted average score of 0.602, ranking first among all evaluated models and outperforming both open-source and proprietary models of substantially larger size. Notably, it secures the highest score on the \emph{Challenging} subset (0.462), demonstrating strong reasoning capabilities and generalization to complex queries. \texttt{Think2SQL-7B} also performs robustly, ranking second overall and surpassing several larger models on both the overall benchmark and the \emph{Challenging} subset. Similarly, \texttt{Think2SQL-3B} attains a weighted average of 0.500, outperforming all models under 10B parameters as well as larger competitors such as \texttt{QwQ-32B} and \texttt{DeepSeek-Llama-70B}. Even the smaller variants—\texttt{Think2SQL-0.5B} and \texttt{Think2SQL-1.5B}—exhibit competitive performance, underscoring the effectiveness of our reasoning-driven training pipeline for enabling high-quality inference in small LLMs.

When directly compared to their non-reasoning counterparts of similar size, the \texttt{Think2SQL} models consistently outperform them. For instance, \texttt{Think2SQL-3B} surpasses \texttt{Qwen2.5-Coder-3B} across all difficulty categories, achieving a weighted average of 0.500 (+0.12). Similarly, \texttt{Think2SQL-7B} outperforms \texttt{Qwen2.5-Coder-7B}, particularly on \emph{Moderate} and \emph{Challenging} instances, with scores of 0.482 vs. 0.388 (+0.09) and 0.385 vs. 0.294 (+0.09), respectively. This improvement trend is more pronounced in smaller models (0.5B, 1.5B, 3B) compared to larger ones (7B, 14B), as illustrated in \cref{fig:model_size_ex_vs_ex}. The enhanced gains in smaller models can be attributed to their limited capacity to internalize complex reasoning patterns during pretraining. 
RLVR techniques like GRPO address this limitation, enabling these models to develop structured reasoning capabilities that are challenging to acquire through supervised fine-tuning alone.

Reasoning capabilities do not always lead to improved performance. For instance, all the distilled DeepSeek models with reasoning perform worse than their non-reasoning counterparts. This suggests that generic reasoning skills are insufficient to solve the Text2SQL task effectively. Instead, these results highlight the importance of task-specific training: models must be explicitly exposed to structured reasoning within the domain to learn how to apply reasoning effectively for the task.

Notably, this underscores the efficacy of task-specific reasoning: medium-to-large models like DeepSeek-Llama (70B) and DeepSeek-Qwen (32B), when relying on general reasoning capabilities, achieve lower performance than our significantly smaller Think2SQL-3B model trained with task-specific reasoning, as detailed in~\cref{tab:bird-dev-analysis}.

\begin{AIbox}{Takeaway 1: Think2SQL}
    Incorporating reasoning significantly enhances model performance on the Text2SQL task, particularly for challenging examples {and small LLMs}. However, general reasoning capabilities alone are insufficient. Task-specific reasoning traces are essential for better performance.
\end{AIbox}

\begin{table}[htbp]
    \setlength{\tabcolsep}{3pt}
    \small
    \centering
    \caption{\textbf{Ablation over different training strategies.} In bold are the best results for each model size. RL$_{EX}$ and RL$_{QATCH}$ are the RL training strategies with EX and QATCH, respectively. SFT$_{NT}$ is the SFT training strategy without reasoning tokens.
        \textcolor{MyGreen}{$\blacktriangle$} indicates the percentage improvement over the corresponding base model.
        Scores reflect EX; the higher the better. All Qwen models are 2.5-Coder-Instruct variants.}
    \label{tab:ablation_training_strategy}
    \begin{tabular}{l@{\hskip 0.1pt}|lll|l}
        \toprule
        \textbf{Model}                    & \textbf{Simple}               & \textbf{Medium}               & \textbf{Challenging}        & \textbf{Weighted AVG}       \\
        \midrule
        \texttt{Qwen-3B}                  & $0.469$                       & $0.267$                       & $0.196$                     & $0.382$                     \\
        \texttt{Qwen-3B-SFT$_{NT}$}       & $0.510$                       & $0.325$                       & $0.224$                     & $0.427$                     \\
        \texttt{Qwen-3B-SFT}              & $0.531$                       & $0.366$                       & $0.301$                     & $0.460$                     \\

        \texttt{Qwen-3B-RL$_{EX}$}        & $0.569$                       & $0.381$                       & $0.273$                     & $0.485$                     \\
        \texttt{Qwen-3B-RL$_{QATCH}$}     & \textbf{0.574}~\improve{22}   & \textbf{0.403}~\improve{51}   & 0.336                       & \textbf{0.500}~\improve{31} \\
        \texttt{Qwen-3B-SFT-RL$_{EX}$}    & $0.560$                       & $0.370$                       & \textbf{0.343}~\improve{75} & $0.482$                     \\
        \texttt{Qwen-3B-SFT-RL$_{QATCH}$} & 0.564                         & 0.372                         & 0.305                       & 0.482                       \\

        \hline
        \texttt{Qwen-7B}                  & $0.548$                       & $0.388$                       & $0.294$                     & $0.476$                     \\
        \texttt{Qwen-7B-SFT$_{NT}$}       & $0.537$                       & $0.310$                       & $0.273$                     & $0.443$                     \\
        \texttt{Qwen-7B-SFT}              & $0.573$                       & $0.40$                        & $0.294$                     & $0.494$                     \\

        \texttt{Qwen-7B-RL$_{EX}$}        & $0.619$                       & $0.463$                       & $0.406$                     & $0.552$                     \\
        \texttt{Qwen-7B-RL$_{QATCH}$}     & $\textbf{0.628}$~\improve{15} & $\textbf{0.482}$~\improve{24} & $0.385$                     & \textbf{0.561}~\improve{18} \\

        \texttt{Qwen-7B-SFT-RL$_{EX}$}    & $0.590$                       & $0.422$                       & $0.343$                     & $0.516$                     \\
        \texttt{Qwen-7B-SFT-RL$_{QATCH}$} & $0.602$                       & $0.442$                       & \textbf{0.427}~\improve{45} & $0.537$                     \\

        \hline
        \texttt{Qwen-14B}                 & 0.610                         & 0.456                         & 0.364                       & $0.541$                     \\
        \texttt{Qwen-14B-RL$_{QATCH}$}    & \textbf{0.654}~\improve{7}    & \textbf{0.543}~\improve{19}   & \textbf{0.462}~\improve{27} & \textbf{0.602}~\improve{11} \\

        \bottomrule
    \end{tabular}
\end{table}

\subsection{Ablation on different training strategies}

\cref{tab:ablation_training_strategy} presents an ablation study comparing different training strategies applied to \texttt{Qwen2.5-Coder} models of two sizes (3B and 7B parameters). Across both model sizes, we observe that applying SFT with reasoning traces consistently improves performance over the base model. Notably, the \texttt{SFT$_{NT}$} variant, which lacks reasoning traces, shows a significant drop in performance compared to the full SFT model, particularly on the more challenging examples.
For the 3B model, the \texttt{SFT} variant outperforms the \texttt{SFT$_{NT}$} of 0.04 points on average, indicating that the reasoning traces are beneficial for the model's performance. Instead, for the 7B model, the \texttt{SFT$_{NT}$} variant performs slightly worse than the base model suggesting possible overfitting to the training data. This highlights the importance of reasoning traces in guiding the model's learning process.
The model trained exclusively with RLVR exhibits further performance gains, particularly on the more challenging subsets. Among the reward functions considered, $R_{\text{QATCH}}$ yields the best results, consistently outperforming $R_{\text{EX}}$ across both the 3B and 7B model sizes. These findings suggest that sparse rewards such as $R_{\text{EX}}$ are less effective in guiding the model's learning process compared to denser, more informative reward signals.

\begin{wraptable}{r}{5.5cm}  
    \centering
    \vspace{-4mm}
    \begin{footnotesize}
        \caption{\textbf{Improvement with R$_{QATCH}$ different model families.}}
        \begin{tabular}{ll}
            \toprule
            Model                            & EX                      \\
            \midrule
            Llama-8B                         & 0.355                   \\
            Llama-8B-RL                      & 0.509~\improvesmall{43} \\
            \midrule
            DeepSeek-Qwen-7B                 & 0.218                   \\
            DeepSeek-Qwen-7B-RL              & 0.307~\improvesmall{41} \\
            \bottomrule
        \end{tabular}
        \label{table:rl_qatch_different_model}
    \end{footnotesize}
\end{wraptable}
For the larger \texttt{Qwen2.5-Coder-7B} model, we observe a clear performance gain over its 3B counterpart across all difficulty levels, confirming that model scale remains a significant factor, particularly when fine-tuned with RLVR via GRPO. The best overall performance is achieved by \texttt{Qwen2.5-Coder-14B-RL$_{QATCH}$}, for which we report results only under the RLVR setup with $R_{\text{QATCH}}$ to avoid redundancy. This model attains the highest average score (0.602) and delivers the strongest results on \emph{Simple}, \emph{Medium}, and \emph{Challenging} examples.

To assess the generalizability of R$_{\text{QATCH}}$-based RLVR, we applied the same training setup to \texttt{Llama-8B} and \texttt{DeepSeek-Qwen-7B} models. Both exhibited similar performance gains of 43\% and 41\%, respectively, as detailed in~\cref{table:rl_qatch_different_model}. These results suggest that the benefits of $R_{\text{QATCH}}$ extend beyond the \texttt{Qwen2.5-Coder} family to other model architectures.


\begin{AIbox}{Takeaway 2: RL vs SFT}
    \textbf{SFT} improves over base models, especially for smaller LLMs. \textbf{RL} improves performance across all difficulty levels, particularly in more complex scenarios.
    \textbf{Dense Rewards} as $R_{\text{QATCH}}$ are more effective than sparse rewards like $R_{\text{EX}}$ for RL.
\end{AIbox}


\begin{table}[htbp]
    \centering
    \caption{\textbf{Analysis robustness for different datasets.} In bold are the best results for each model size. The RL$_{EX}$ and RL$_{QATCH}$ are the RLVR training strategies with $R_{\text{EX}}$ and $R_{\text{QATCH}}$, respectively. \textcolor{MyGreen}{$\blacktriangle$} indicates the percentage improvement over the corresponding base model.
        Scores reflect EX; higher is better. All Qwen models are 2.5-Coder-Instruct variants.}
    \label{tab:ablation_diff_tables}
    \begin{tabular}{lllll}
        \toprule
        \textbf{Model}                    & \textbf{Spider}            & \textbf{Spider-Syn}         & \textbf{Spider-DK}            & \textbf{KaggleDBQA}         \\
        \midrule
        \texttt{Qwen-3B}                  & $0.725$                    & $0.632$                     & $0.602$                       & 0.252                       \\
        \texttt{Qwen-3B-SFT}              & $0.770$                    & $0.718$                     & $0.634$                       & 0.312                       \\ 
        \texttt{Qwen-3B-RL$_{QATCH}$}     & $0.770$                    & $0.717$                     & $\textbf{0.680}$~\improve{13} & 0.360                       \\ 
        \texttt{Qwen-3B-SFT-RL$_{QATCH}$} & \textbf{0.772}~\improve{6} & \textbf{0.725}~\improve{15} & 0.655                         & \textbf{0.387}~\improve{54} \\
        \hline
        \texttt{Qwen-7B}                  & $0.776$                    & $0.703$                     & $0.652$                       & 0.247                       \\ 
        \texttt{Qwen-7B-SFT}              & $0.799$                    & $0.774$                     & $0.652$                       & 0.380                       \\ 
        \texttt{Qwen-7B-RL$_{QATCH}$}     & $0.822$                    & $0.769$                     & \textbf{0.731}~\improve{12}   & 0.436                       \\
        \texttt{Qwen-7B-SFT-RL$_{QATCH}$} & \textbf{0.826}~\improve{6} & \textbf{0.794}~\improve{13} & 0.687                         & \textbf{0.441}~\improve{79} \\  
        \bottomrule
    \end{tabular}
\end{table}
\subsection{Reasoning robustness on different datasets}
\cref{tab:ablation_diff_tables} reports EX of various training strategies across different dataset variants: the original Spider dataset, its synonym-augmented version (Spider-Syn), the domain-knowledge variant (Spider-DK), and the more challenging KaggleDBQA. All training strategies consistently outperform the base \texttt{Qwen2.5-Coder} models for both the 3B and 7B sizes, confirming the generalizability of our training strategies. Specifically, the 7B size has gained 6-79\% improvement range over the base model, while the 3B size has gained 6-54\% improvement range.

The combined \texttt{SFT-RL$_{QATCH}$} strategy achieves the best overall performance, which yields the highest accuracy on all the datasets besides Spider-DK, demonstrating stronger generalization when reasoning traces are incorporated during training. However, on Spider-DK, which demands deeper domain knowledge, pure RL appears to offer an advantage—\texttt{RL$_{QATCH}$} outperforms all other strategies on this subset for both model sizes.

Notably, the combined strategy of leveraging high-quality reasoning traces from larger models and incorporating them into smaller models via SFT appears to enhance performance. Analyzing the inference reasoning traces, we empirically see that smaller models inherit the structured cognitive behaviors developed by their larger counterparts~\citep{ranaldi2024self,gandhi2025cognitive}, probably facilitating more effective reasoning. An example of two different reasoning traces can be seen in \cref{sec:example-inference-reasoning}.

\begin{AIbox}{Takeaway 3: Generalization with SFT + RL}
    Combining SFT and RL yields the most generalizable models, excelling across diverse datasets. This suggests that integrating reasoning traces with SFT enables models to better adapt to complex and unseen scenarios.
\end{AIbox}


\section{Related Work}

Recent Text2SQL approaches apply a variety of reasoning-centric strategies to
decompose the complex SQL generation pipeline into step-by-step reasoning processes~\cite{zhang2023act,pourreza2023din}. 
They adopt two main techniques:
(1) LLM Supervised Fine-Tuning~\cite{lyu2025sql,wang2019ratSql},
(2) LLM prompting techniques, such as Chain-of-Thought and Few-Shot Learning~\cite{li2024dawn,talaei2024chessSql}. Concurrently with our work, several very recent studies have started to explore Reinforcement Learning~\cite{NGUYEN2025100135,zhai2025excot,pourreza2025reasoning}.
Reinforcement Learning with Verifiable Reward has proved to be particularly effective in complex reasoning tasks, such as mathematical reasoning and coding. However, the combined use of SFT with reasoning traces and RLVR in Text2SQL remains unexplored.  
To bridge the gap, this paper investigates the impact of reasoning and RLVR with partial scoring rewards on Text2SQL performance.



\section{Conclusions and Limitations}
\label{sec:concl}

Our work demonstrates that targeted {reasoning reinforcement} is a decisive ingredient for Text2SQL. While traditional fine-tuning  yields only marginal gains, injecting structured reasoning traces during supervised fine-tuning (SFT) markedly increases execution accuracy, especially for models below 10B parameters.
Reinforcement Learning with Verifiable Reward (RLVR) further improves performance across all model sizes; dense, partially graded rewards provide informative gradients even when the generated SQL is incorrect, in contrast with binary execution accuracy rewards.

With RLVR we obtain our 14B parameter \texttt{Think2SQL} model, which reaches a weighted execution accuracy of 0.602 on the BIRD benchmark, outperforming proprietary GPT-4o and open-source systems 5–45x larger.
These results confirm that targeted reasoning supervision can offset two orders of magnitude in parameter count.

\begin{table}[h]
\centering
  \caption{Key limitations of the present study.}\centering
  \small
  \begin{tabular}{@{}p{2.5cm}p{4.7cm}p{6.2cm}@{}}
    \toprule
    \textbf{Aspect}        & \textbf{Current scope}                                                       & \textbf{Implication}                                                                                 \\
    \midrule
    Schema linking         & Gold, question-specific tables provided in-context.                          & Real deployments must retrieve tables automatically; linking errors may reduce accuracy.             \\
    Inference heuristics   & No self-reflection, self-consistency, or ensembling.                         & Single-pass results underestimate the attainable ceiling with lightweight inference-time heuristics. \\
    Reward design          & Rewards linearly rescaled to $[0,1]$ with fixed weights (0.85/ 0.10 / 0.05). & Adaptive weighting may speed learning.                                                               \\
    RLVR recipe            & Emerging methods such as DAPO, zero-RL, 1-shot RL not explored.              & Newer value-free variants could reduce compute and improve stability.                                \\
    Reasoning-trace source & SFT traces generated by DeepSeek-R1.                                         & Their stylistic bias may affect outcomes; human or mixed-initiative traces could change results.     \\
    \bottomrule
  \end{tabular}
  \label{tab:limitations}
\end{table}

\textbf{Limitations.} \cref{tab:limitations} summarizes the principal limitations under which our study was conducted and outlines their practical implications.

\acksection
This project was provided with computer and storage resources by GENCI at IDRIS, thanks to grants 2025-AD010616649 and 2025-AD010616180 on the supercomputer Jean Zay, using the H100 and A100 partitions.

\bibliographystyle{abbrvnat}
\bibliography{biblio}
\appendix
\newpage
\section{Fine-Tuning Language Models}
\subsection{Supervised Fine-Tuning}
\label{sec:finetuning}
Supervised Fine-Tuning (SFT) adapts a pretrained language model $\pi_{\mbtheta}$ to a distribution $\cP$ of sequences that reflect desired linguistic or task-specific behavior.
Let $\mbz = (z_1, z_2, \ldots, z_T)$ denote a token sequence drawn from $\mbz \sim \cP$.
The SFT objective maximizes the likelihood of sequences under $\pi_{\mbtheta}$, which corresponds to minimizing the expected negative log-likelihood:
\begin{equation}
    \label{eq:nll-loss-sft-full}
    \cL_{\text{full}}(\mbtheta) = - \E_{\mbz \sim \cP} \left[ \sum_{t=1}^{T} \log \pi_{\mbtheta}(z_t \mid \mbz_{<t}) \right].
\end{equation}

For tasks with an explicit decomposition into an input segment and a target segment—such as QA, summarization, or assistant-style dialogue—the data distribution consists of pairs $(\mbx, \mby) \sim \cP$, where $\mbx = (x_1, \ldots, x_n)$ is the conditioning prompt and $\mby = (y_1, \ldots, y_m)$ is the supervised output. In such settings, the model conditions on $\mbx$ and predicts the continuation $\mby$, with the loss computed only over the target tokens:
\begin{equation}
    \label{eq:nll-loss-sft-response}
    \cL_{\text{cond}}(\mbtheta) = - \E_{(\mbx, \mby) \sim \cP} \left[ \sum_{t=1}^{m} \log \pi_{\mbtheta}(y_t \mid \mbx \Vert \mby_{<t}) \right]\,,
\end{equation}
where $\Vert$ denotes the concatenation operator.
This alternative objective is often preferred in practice, as it allows for more efficient training by focusing on the relevant output tokens and ignoring the input tokens \citep{chiang2023vicuna,yu-etal-2024-lions,wang-etal-2023-self-instruct}.
More recently, \citet{shi2024instruction} showed that models trained with the SFT objective in \cref{eq:nll-loss-sft-full} can be superior to \cref{eq:nll-loss-sft-response} when the target sequence is significantly shorter than the input sequence.
In the case of distillation of reasoning models, the output sequence will be considerebly longer than the input sequence, and the SFT objective in \cref{eq:nll-loss-sft-response} is preferred.
Finally, the expectations in \cref{eq:nll-loss-sft-full} and \cref{eq:nll-loss-sft-response} are approximated by empirical means over a finite dataset $\cD = \{\mbz_i\}_{i=1}^N$ or $\cD = \{(\mbx_i, \mby_i)\}_{i=1}^N$ consisting of $N$ training examples.
The resulting objective is optimized via standard stochastic gradient descent or its variants \cite{robbins1951stochastic,kingma2014adam}.

\subsection{Group-Relative Policy Optimization}
\label{sec:grpo}
Group-Relative Policy Optimization (GRPO)~\cite{shao2024deepseekmath} has been recently introduced as a value-free alternative to Proximal Policy Optimization (PPO)~\cite{schulman2017PPO} for fine-tuning language models.

Let $\mbx \sim \cX$ denote a prompt drawn from a distribution over conditioning inputs, and let $\{\mby_i\}_{i=1}^G \sim \pi_{\mbtheta_{\text{old}}}(\cdot \mid \mbx)$ be $G$ response sequences generated by the frozen reference policy $\pi_{\mbtheta_{\text{old}}}$. Each response $\mby_i = (y_{i,1}, \ldots, y_{i,T_i})$ is assigned a scalar reward $R_i \in \bbR$ computed via a reward model.

The group-relative advantage $A_i$ for the $i$-th response is defined by normalizing the reward distribution over the group:
\begin{equation}
    \label{eq:group-advantage}
    A_i = \frac{R_i - \E[R_j]}{\sqrt{\mathbb{V}[R_j]}}, \qquad j \in \{1, \ldots, G\},
\end{equation}
where $\E[R_j]$ and $\mathbb{V}[R_j]$ are the mean and variance of the rewards for the group of responses, respectively.
For each token position $t$ in response $\mby_i$, define the state as $\mbs_{i,t} = \mbx \Vert \mby_{i,<t}$, and the token-level probability ratio as
$$
    p_{i,t}(\mbtheta) = \frac{\pi_{\mbtheta}(y_{i,t} \mid \mbs_{i,t})}{\pi_{\mbtheta_{\text{old}}}(y_{i,t} \mid \mbs_{i,t})}.
$$
The GRPO training objective maximizes a clipped surrogate loss penalized by the KL divergence from the reference policy:
\begin{equation}
    \label{eq:grpo-surrogate-loss}
    \begin{aligned}
        \cL_{\text{GRPO}}(\mbtheta) =
        \E
        \left[
            \frac{1}{G} \sum_{i=1}^{G} \frac{1}{T_i} \sum_{t=1}^{T_i}
            \min\left(
            p_{i,t}(\mbtheta) A_i,\;
            \text{clip}\big(p_{i,t}(\mbtheta), 1 - \epsilon, 1 + \epsilon\big) A_i
            \right)
            - \beta \KL{\pi_{\mbtheta}}{\pi_{\mbtheta_{\text{ref}}}}
            \right],
    \end{aligned}
\end{equation}
where the expectation is taken over the prompt distribution $\mbx \in \cX$, and the responses $\{\mby_i\}_{i=1}^G$ generated by the frozen policy $\pi_{\mbtheta_{\text{old}}}$.
Additionally, $\epsilon$ is set to be the clipping parameter and $\beta$ controls the \gls{KL} regularization by penalizing models that deviate from the reference policy $\pi_{\mbtheta_{\text{ref}}}$ (which is typically the initial pretrained model).

\subsection{Rule-based Reward Modeling}
Reward modeling is central to reinforcement learning with language models, as it defines the optimization signal guiding the policy $\pi_{\mbtheta}$.
Learned neural reward models are commonly employed to approximate human preferences or task-specific goals.
However, they often suffer from distributional mismatch, reward hacking, and spurious correlations \citep{gao2023rewardhack,weng2024rewardhack,everitt2021rewardhack}.
These effects arise when the model exploits imperfections in the reward predictor, leading to high-reward outputs that do not correspond to true task success.

An alternative is to design rule-based reward models, which define deterministic mappings from model outputs to scalar reward values via explicit criteria.
In the context of coding, for instance, a reward function $R: \mby \mapsto [0,1]$ can be constructed by executing the generated code $\mby$ against a test suite and returning the fraction of passed unit tests.
Such rule-based models directly encode correctness and task satisfaction, avoiding pathologies introduced by learned approximators.

Formally, let $\mbx \sim \cX$ denote the input (e.g., a natural language instruction), and $\mby \sim \pi_{\theta_{\text{old}}}(\cdot \mid \mbx)$ a candidate response.
The reward function $R(\mbx, \mby) \in \bbR$ is defined deterministically via evaluation procedures specified a priori.
These functions are task-dependent and vary across application domains. The resulting reward is used to construct advantage estimates, as in GRPO.

A well-known limitation of rule-based reward models is the sparsity of the reward signal.
In many structured tasks, the reward $R(\mbx, \mby)$ may remain zero across most model outputs and attain nonzero values only when the generation exactly satisfies task constraints.
This sparsity complicates credit assignment during training and may impair exploration in RL-based optimization.
Techniques such as reward shaping, curriculum learning, or relaxed matching criteria are sometimes introduced to mitigate this issue~\cite{ng1999policy, sutton1998reinforcement, gao2023rewardhack,narvekar2020curriculum}.
Nonetheless, provided that the policy starts from a sufficiently strong pretrained model, this approach has been successfully adopted in multiple recent frameworks across general and specialized RLHF pipelines~\cite{deepseekai2025deepseekr1,team2025kimi,yu2025dapoopensourcellmreinforcement}, and has been particularly effective in settings where ground truth verification criteria exist, such as program synthesis~\cite{le2022coderl,gehring2024rlef,chen2023teaching}.

\section{Prompt Synthetic data annotation}
\label{sec:prompt-synthetic-data-annotation}

\begin{figure*}[htbp]
    \centering
    \begin{tcolorbox}[title=Prompt Synthetic data annotation, promptstyle]
        \lstset{
            basicstyle=\normalfont\sffamily\footnotesize,
            breaklines=true,
            frame=none,
            columns=fullflexible,
        }
        Answer the following question with the SQL code. Use the piece of evidence and base your answer on the database schema.\\
        Given the question, the evidence and the database schema, return in the answer tags only the SQL script that addresses the question.

        Question:\\
        \texttt{<question>}

        Evidence:\\
        \texttt{<evidence>}

        Database Schema:\\
        \texttt{<schema>}

    \end{tcolorbox}
    \caption{Prompt used for the synthetic data annotation. \texttt{<question>}, \texttt{<evidence>}, and \texttt{<schema>} are placeholders for the actual question, evidence, and database schema, respectively. The model is expected to generate a SQL code snippet that answers the question based on the provided evidence and schema.}
    \label{fig:sft_prompt}
\end{figure*}

The template depicted in Fig.~\ref{fig:sft_prompt} is used in our synthetic–data pipeline.
By isolating the \verb|<question>|, the clarifying \verb|<evidence>|, and the relevant \verb|<schema>|, it forces the model to reason internally and then produce \emph{only} the final SQL enclosed in \verb|<answer>| tags.
This strict I/O contract both curbs hallucinations and makes automatic execution-based grading trivial, yielding high-quality self-annotated examples that we later exploit for supervised fine-tuning and RLVR.

\section{Prompt used for training}
\label{sec:prompt-training}

\begin{figure*}[htbp]
    \centering
    \begin{tcolorbox}[title=Complete training/inference prompt., promptstyle]
        \lstset{
            basicstyle=\normalfont\sffamily\footnotesize,
            breaklines=true,
            frame=none,
            columns=fullflexible,
        }
        \texttt{<|im\_start|>system}

        You are a helpful AI Assistant that provides well-reasoned and detailed responses. You first think about the reasoning process as an internal monologue and then provide the user with the answer. Respond in the following format: \texttt{<think>}

        ...

        \texttt{</think>}

        \texttt{<answer>}

        ...

        \texttt{</answer><|im\_end|>}

        \texttt{<|im\_start|>user}

        Answer the following question with the SQL code. Use the piece of evidence and base your answer on the database schema. Given the question, the evidence and the database schema, return in the \texttt{<answer>} tags only the SQL script that addresses the question.

        \textbf{Question}:

        Calculate the average of 2020's population in each zip code.

        \textbf{Evidence}:

        average of 2020 population in each zip code refers to Divide (Sum(population\_2020), Count(zip\_code))

        \textbf{Database Schema}:

        CREATE TABLE zip\_data

        (

        zip\_code                         INTEGER

        primary key,

        city                             TEXT,

        state                            TEXT,

        multi\_county                     TEXT,

        "1st\_quarter\_payroll"            INTEGER,

        ...

        foreign key (state) references state(abbreviation),

        foreign key (CBSA) references CBSA(CBSA)

        )

        Return only the SQL script enclosed in \texttt{<answer>} tags.\texttt{<|im\_end|>}

        \texttt{<|im\_start|>assistant}

    \end{tcolorbox}
    \caption{Complete prompt tokenized for the model Qwen2.5-Coder-Instruct used during training and inference. During SFT the prompt is extended with the reasoning process, and the answer.}
    \label{fig:training_prompt}
\end{figure*}

The template in Fig.~\ref{fig:training_prompt} encodes the supervision signal in three blocks.
A \texttt{system} message first primes Qwen2.5-Coder-Instruct to reason privately between \verb|<think>|…\verb|</think>| tags and to surface only the executable SQL inside \verb|<answer>| tags.
The \texttt{user} turn bundles the natural-language question, schema-clarifying evidence, and the database schema itself, guaranteeing that every token required for generation is present in a single prompt.
During supervised fine-tuning we append an oracle reasoning trace plus the target query, so the model is teacher-forced on both latent reasoning and final output; at inference the identical skeleton is reused, maintaining a contract that simplifies reward computation in RLVR.

\section{Rewards for RLVR}
\label{sec:rewards}
In reinforcement learning, reward signals are crucial for guiding the model's learning process~\citep{gao2023rewardhack,weng2024rewardhack,everitt2021rewardhack}. Execution accuracy, the primary reward for Text2SQL, measures the correctness of generated SQL by comparing it to the ground truth. However, its binary nature poses challenges for RL optimization, especially for smaller LLMs, as rewards often remain zero unless the SQL is exactly correct.
To address this limitation, we integrate QATCH~\citep{papicchio2023qatch,papicchio2025qatch}, an advanced benchmarking framework designed for the automated evaluation of Text2SQL tasks.
For the purposes of this study, we employed three primary QATCH metrics: \textit{Cell Precision}, \textit{Cell Recall}, and \textit{Tuple Cardinality}.

To encourage the model's reasoning process, we introduce the \textit{Format reward} that evaluates the appropriate use of reasoning tags~\cite{deepseekai2025deepseekr1}. Additionally, to mitigate reward hacking, the \textit{Tag count reward} penalize the reward when reasoning tokens are redundantly or excessively repeated within the reasoning trace.

Let $\cT$ and $\cT_{\text{pred}}$ denote the execution results of the target SQL query and the predicted SQL query, respectively, each represented as a set of tuples, where each tuple comprises a set of cell values. The reward signals utilized in this study are outlined below:

\stitle{Execution Accuracy (EX).} EX~\citep{yu2018spider,li2024bird} evaluates whether the execution of the target SQL query matches the execution of the predicted SQL query. It is defined as:
\begin{equation}
    R_{\text{EX}} =
    \begin{cases}
        1 & \text{if } \cT = \cT_{\text{pred}} \\
        0 & \text{otherwise}
    \end{cases}, \quad R_{\text{EX}} \in \{0, 1\}
\end{equation}

This metric provides a binary reward, assigning a full score only when the two execution results match exactly, row by row. While straightforward and reliable, execution accuracy does not account for partially correct results, which can hinder the learning process in RL.

\stitle{Cell Precision (CP)}. CP is the fraction of table cells in $\cT_{\text{pred}}$ that are in the target $\cT$. The higher the score, the more predicted cells are in the target.
\begin{equation}
    R_{\text{CP}} =
    \frac{
        |\{ cells \mid cells \in \cT \cap \cT_{\text{pred}} \}|
    }{
        |\{cells \mid cells \in \cT\}|
    }, \quad R_{\text{CP}} \in [0, 1]
\end{equation}
This metric allows for partial credit when the predicted SQL query execution contains some requested cells but also includes incorrect ones. Considering the target query
\lstinline[style=SQL]!SELECT Name FROM Player;!
and the predicted query
\lstinline[style=SQL]|SELECT Name,|
{\small\ttfamily{\textcolor{red}{\underline{Surname}}}}
\lstinline[style=SQL]|FROM Player;|
in this case CP is $0.5$ because the predicted SQL query execution contains the correct cells from the column Name and incorrect ones from Surname. However, it does not consider whether all the requested cells in $\cT$ are present in the SQL query - measured by \textit{Cell Recall}. It is worth noticing that when EX is $1$ also CP is $1$.

\stitle{Cell Recall (CR)}. CR is the fraction of table cells in $\cT$ that are present in $\cT_{\text{pred}}$. The higher the score, the more target cells are included in the prediction.
\begin{equation}
    R_{\text{CR}} =
    \frac{
        |\{ cells \mid cells \in \cT \cap \cT_{\text{pred}} \}|
    }{
        |\{cells \mid cells \in \cT_{\text{pred}}\}|
    }, \quad R_{\text{CR}} \in [0, 1]
\end{equation}
This metric allows a partial reward in case the predicted SQL query does not contain all the requested cell in the target query. Considering the target query
\lstinline[style=SQL]|SELECT Name, Surname FROM Player;|
and the predicted query
\lstinline[style=SQL]|SELECT Name,|
{\small\ttfamily{\textcolor{red}{\sout{Surname}}}}
\lstinline[style=SQL]|FROM Player;|
then CP is $1$ but CR is $0.5$ because the predicted SQL query execution contains the correct cells from the column Name but not from Surname. It is worth noticing that when EX is $1$ CR is $1$ as well.

\stitle{Tuple Cardinality (TC)}. TC is defined as the ratio between the number of tuples in $\cT_{\text{pred}}$ and the number of tuples in $\cT$. The $\min$ function is used to ensure  $\text{TC} \in [0,1]$. TC captures output cardinality only, ignoring schema and cell values. Thus, it should be considered alongside CP and CR for a fuller view of model performance. The TC reward is defined as:
\begin{equation}
    R_{\text{TC}} = \min\left( \frac{|\cT|}{|\cT_{\text{pred}}|}, \frac{|\cT_{\text{pred}}|}{|\cT|} \right), \quad R_{\text{TC}} \in [0, 1]
\end{equation}
This metric is necessary because CP and CR are computed based on the intersection of cell values, which may overlook differences in output size when the number of cells is not critical. For example, consider the target query
\lstinline[style=SQL]!SELECT DISTINCT Name FROM Player;!
and the predicted query
\lstinline[style=SQL]!SELECT!
{\small\ttfamily{\textcolor{red}{\sout{DISTINCT}}}}
\lstinline[style=SQL]!Name FROM Player;!.
In this case, both CP and CR equal $1$ due to identical cell values, yet the cardinality of the cells is different. Thus, the TC metric is essential for capturing this discrepancy.

\stitle{Format Reward (FR).} The FR~\cite{deepseekai2025deepseekr1} incentivizes the model to adhere to a predefined output structure, such as the use of \verb!<think>!
and \verb!<answer>! tags.
\begin{equation}
    R_{\text{FR}} = \begin{cases}
        1 & \text{if } \pi_{\mbtheta}(\mbx) \text{ matches } \texttt{{\small <think/>.*?</think>s*<answer>.*?</answer>}} \\
        0 & \text{otherwise}
    \end{cases}, \quad R_{\text{FR}} \in \{0, 1\}
\end{equation}
This is a sparse reward that activates only when both the opening and closing tags for reasoning and answers are correctly positioned. The reward value is 1 if the tags are correctly formatted; otherwise, it is 0.

\stitle{Tag Count Reward (TCR).} To address reward hacking, where reasoning traces include unnecessary or excessive tags, we introduce the TCR. This reward penalizes the model for generating reasoning traces with redundant tags. The reward is 1 if each tag appears exactly once in the reasoning trace, and decreases proportionally for each redundant or missing tag. Considering $t \in \{\texttt{<think>}, \texttt{</think>}, \texttt{<answer>}, \texttt{</answer>}\}$:

\begin{equation}
    R_{\text{TCR}} = 0.25 \cdot \sum_{t} \mathds{1}(Count(\pi_{\mbtheta}(\mbx), t) = 1),
    \quad R_{\text{TCR}} \in \{0, 0.25, 0.50, 0.75, 1.0\}
\end{equation}

\stitle{Final Reward.} The final reward signal is computed as a weighted sum of the individual rewards. The weights were carefully chosen to ensure a balanced contribution from each reward component while maintaining a total score of 1. This design prevents training instability caused by excessively high rewards.
In addition, since CP, CR, and TC must be seen together to provide a complete picture of the model's performance, we decided to use the average of these three metrics $R_{\text{QATCH}}= \E[R_{\text{CP}}, R_{\text{CR}}, R_{\text{TC}}]$.

Let $R_{\text{text2SQL}}$ be $R_{\text{EX}}$ or $R_{\text{QATCH}}$, with the combination left for future study.
The final reward is computed as follows:
\begin{equation}
    R = 0.85 \cdot R_{\text{text2SQL}} + 0.10 \cdot R_{\text{Format}} + 0.05 \cdot R_{\text{Tag Count}}, \qquad R \in [0,1]
\end{equation}
These weights are selected to ensure that the execution accuracy and QATCH metrics are the primary focus of the training process, while still encouraging the model to produce well-structured outputs with appropriate reasoning traces.

\section{Reasoning trace examples}
\label{sec:example-inference-reasoning}

\begin{figure*}[htbp]
    \centering
    \begin{tcolorbox}[title=RL generated trace]
        \lstset{
            basicstyle=\normalfont\sffamily\tiny,
            breaklines=true,
            frame=none,
            columns=fullflexible,
        }
        \begin{verbatim}        
<think>
To solve this problem, we need to follow these steps:
1. Identify the patients who have a body mass index (BMI) recorded
in the observations table.
2. Filter the records to find the lowest BMI value.
3. Retrieve the full name (first and last) of the patient with the
lowest BMI.

The evidence provided indicates that the description we are looking for 
is "Body Mass Index" and the units should be 'kg/m2'. 
We also need to find the minimum value of the BMI.

We can use a subquery to find the minimum value of BMI and then
join this with the patients table to get the full name of the patient.
</think>
<answer>
SELECT p.first, p.last FROM patients p
JOIN (
    SELECT PATIENT, MIN(VALUE) AS min_bmi
    FROM observations
    WHERE DESCRIPTION = 'Body Mass Index' AND UNITS = 'kg/m2'
    GROUP BY PATIENT
) o ON p.patient = o.PATIENT
WHERE o.min_bmi IN (
    SELECT MIN(VALUE)
    FROM observations
    WHERE DESCRIPTION = 'Body Mass Index' AND UNITS = 'kg/m2'
)
</answer>

\end{verbatim}

    \end{tcolorbox}
    \caption{Concise reasoning trace produced \emph{after \textbf{RL}-only training}.
        The model emits just a few task-oriented steps before the final SQL, mirroring the sparse,
        verifiable reward that values correctness over explanatory detail.}
    \label{fig:RL_trace}
\end{figure*}

\begin{figure*}[htbp]
    \centering
    \begin{tcolorbox}[title=RL generated trace]
        \lstset{
            basicstyle=\normalfont\sffamily\tiny,
            breaklines=true,
            frame=none,
            columns=fullflexible,
        }
        \begin{verbatim}      
<think>
Okay, let's see. The question is asking for the names of products where
there have been over 3 price changes. The evidence says that "over 3
price changes" means the count of ListPrice greater than 3. 
Hmm, maybe that's a misunderstanding. 
Wait, count(ListPrice)>3 would mean more than three ListPrices,
but that doesn't make sense. Maybe it's actually referring to the
number of entries in ProductListPriceHistory where StartDate is not 
null and ListPrice is not null. Because if a product has multiple
entries, each with different prices, that counts as a change.

So, the actual approach would be:
count the distinct Products where there are entries in 
ProductListPriceHistory with ListPrice higher than a certain threshold. 
But the evidence says over 3 refers to count(ListPrice)>3, which
confuses me. Maybe the intended solution counts distinct [...]
 But that might be more complex. For example:

SELECT DISTINCT p.Name
FROM "Product" p
JOIN ProductListPriceHistory h ON p.ProductID = h.ProductID
WHERE h.ListPrice > h.StartDate AND COUNT(DISTINCT h.StartDate) > 3;

But wait, that's not right.
Because if two entries have the same StartDate (same ProductID),
that counts as one. For example, [...]
Find distinct products that have entries with ListPrice greater than 
some threshold (say 3). For example:

SELECT DISTINCT Name
FROM Product
WHERE ProductID IN (
    SELECT DISTINCT ProductID
    FROM ProductListPriceHistory
    WHERE ListPrice > 3
);

[...]

Alternatively, the user might have intended that for each Product 
in ProductListPriceHistory, if there's an entry with ListPrice >3, 
then include that product. So the answer is selecting distinct names
from product, joining with ProductListPriceHistory where ListPrice >3.
</think>

<answer>
SELECT DISTINCT Name 
FROM Product 
JOIN ProductListPriceHistory ON
            Product.ProductID = ProductListPriceHistory.ProductID 
WHERE ProductListPriceHistory.ListPrice > 3;
</answer>
\end{verbatim}

    \end{tcolorbox}
    \caption{Verbose reasoning trace produced \emph{after \textbf{SFT}\,+\,\textbf{RLVR} training}.
        The policy retains the multi-step chain-of-thought distilled during SFT, so the trace
        includes hypothesis-checking and clarification before the SQL answer.}
    \label{fig:SFTRL_trace}
\end{figure*}

The reasoning trace in Figure~\ref{fig:RL_trace} is roughly one-fourth of the length of the full trace in Figure~\ref{fig:SFTRL_trace},
because pure RLVR encourages the agent to minimize tokens that do not directly affect the
reward, whereas the SFT phase seeds longer ``show-your-work’’ patterns that persist after
subsequent RLVR fine-tuning. The latter is driven by the detailed reasoning traces originally produced by DeepSeek and used in the SFT step.

This brevity does not hurt in-domain accuracy: the RL-only model attains the best
execution accuracy on the BIRD development set.
However, adding SFT traces before RLVR yields the most robust models overall, achieving the
highest average accuracy on cross-domain benchmarks such as Spider-Syn, Spider-DK, and
KaggleDBQA, indicating better generalization beyond BIRD.

\end{document}